\documentclass[runningheads]{llncs}

\usepackage{eccv}

\usepackage{eccvabbrv}

\usepackage{graphicx}
\usepackage{booktabs}

\usepackage[accsupp]{axessibility}  
\usepackage{multirow} 
\usepackage{wrapfig}

\usepackage[pagebackref,breaklinks,colorlinks,citecolor=eccvblue]{hyperref}

\usepackage{orcidlink}

\begin{document}

\title{Self-Localized Collaborative Perception} 


\author{Zhenyang Ni\inst{1}\and
Zixing Lei\inst{1} \and
Yifan Lu\inst{1} \and
Dingju Wang\inst{1} \and
Chen Feng\inst{2} \and
Yanfeng Wang \inst{1,3} \and
Siheng Chen \inst{1,3*}
}

\authorrunning{Z. Ni, Z. Lei et al.}

\institute{Cooperative Medianet Innovation Center , Shanghai Jiao Tong University, China. \and
\email{\{0107nzy,chezacarss,yifan\_lu,wangdingju,wangyanfeng622,sihengc\}@sjtu.edu.com}\\
New York University, USA. \and
\email{cfeng@nyu.edu} \\
Shanghai AI laboratory, China.
}

\maketitle

\begin{abstract}
Collaborative perception has garnered considerable attention due to its capacity to address several inherent challenges in single-agent perception, including occlusion and out-of-range issues. However, existing collaborative perception systems heavily rely on precise localization systems to establish a consistent spatial coordinate system between agents. This reliance makes them susceptible to large pose errors or malicious attacks, resulting in substantial reductions in perception performance. To address this, we propose~$\mathtt{CoBEVGlue}$, a novel self-localized collaborative perception system, which achieves more holistic and robust collaboration without using an external localization system. The core of~$\mathtt{CoBEVGlue}$ is a novel spatial alignment module, which provides the relative poses between agents
by effectively matching co-visible objects across agents. We validate our method on both real-world and simulated datasets. The results show that i) $\mathtt{CoBEVGlue}$ achieves state-of-the-art detection performance under arbitrary localization noises and attacks; and ii) the spatial alignment module can seamlessly integrate with a majority of previous methods, enhancing their performance by an average of $57.7\%$. Code is available at \href{https://github.com/VincentNi0107/CoBEVGlue}{https://github.com/VincentNi0107/CoBEVGlue}.
\end{abstract}

\section{Introduction}
\label{sec:intro}

\begin{figure}[h]
\hsize=\textwidth 
\centering
\includegraphics[width=\textwidth]{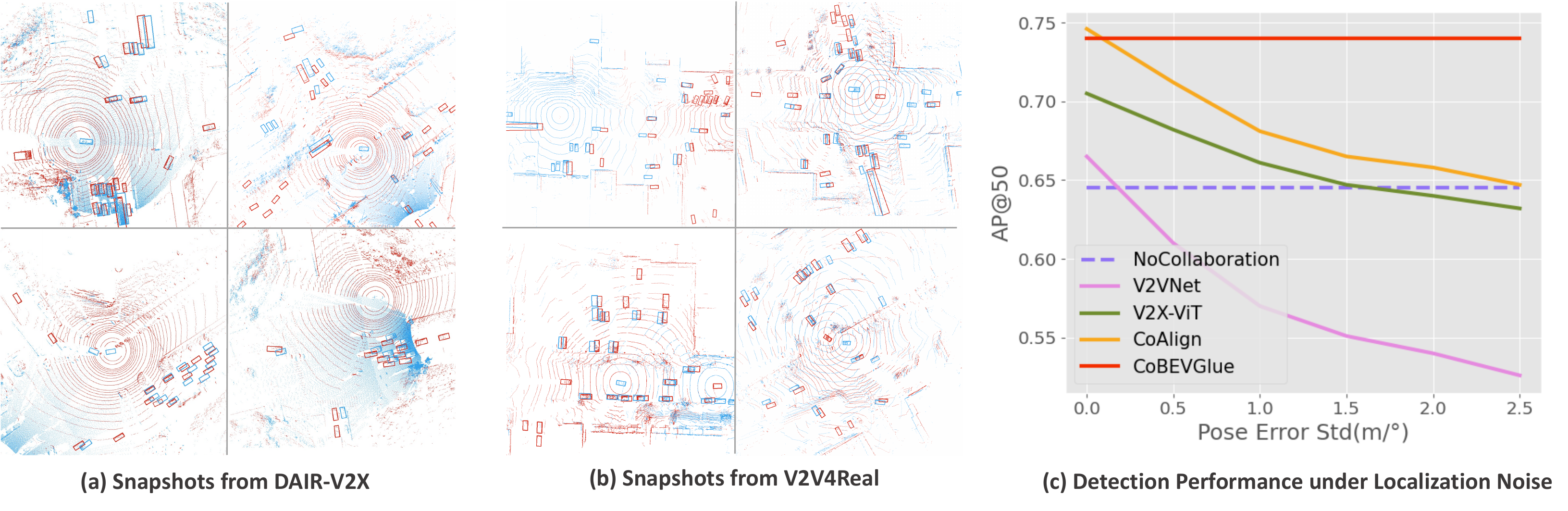}
\caption{ {Noise is pervasive in external localization systems, leading to substantial reductions on the detection performance of collaborative perception systems.} (a) and (b) show snapshots from the real-world collaborative perception dataset DAIR-V2X \cite{yu2022dair} and V2V4Real\cite{xu2023v2v4real}. \textcolor{red}{Red} denotes point cloud and ground truth bounding boxes from the ego agent and \textcolor{blue}{blue} belongs to the collaborator. The point cloud and bounding boxes from collaborator are transformed to the coordinate system of ego agent with ground truth poses. Despite resource-intensive offline calibration efforts, the ground truth localization error persists at the meter-level. (c) shows that current collaborative perception systems fail to transcend the no collaboration baseline under large localization noise on DAIR-V2X.{ In comparison, our $\mathtt{CoBEVGlue}$ achieves state-of-art detection performance when localization noise exist, performing comparably to systems relying on precise localization information.} }
\label{fig:teaser}
\end{figure}

Accurate perception is essential for the navigation and safety of autonomous vehicles \cite{liu2023towards,meng2023hydro}. Despite advancements facilitated by large-scale datasets~\cite{caesar2020nuscenes,sun2020waymo}, and powerful models~\cite{lang2019pointpillars,yin2021center}, single-agent perception is inherently limited by occlusions and long-range issues \cite{wang2020v2vnet}, which could lead to catastrophic consequences \cite{zhang2021safe}. Leveraging modern communication technologies, current research in collaborative perception \cite{li2021learning,hu2022where2comm,xu2022v2x,wang2020v2vnet} enables the sharing of perceptual information among multiple agents, fundamentally improving the perception performance. Fueled by the advent of high-quality datasets \cite{xu2022opv2v,yu2022dair,xu2023v2v4real,li2022v2x} and innovative collaborative techniques \cite{liu2020when2com,liu2020who2com,zhou2022multi,arnold2020cooperative,xu2022cobevt,xiang2023hm}, collaborative perception systems have the potential to improve the safety of transportation networks significantly.

In this emerging field of collaborative perception, most prevailing works \cite{wang2020v2vnet,li2021learning,hu2022where2comm} make an oversimplified assumption: the global localization system, typically GPS or SLAM,  employed by each agent is precise enough to establish a consistent spatial coordinate system for collaboration. However, snapshots from real-world collaborative perception datasets V2V4Real \cite{xu2023v2v4real} and DAIR-V2X \cite{yu2022dair} show that the ground truth localization is still noisy even after meticulous and resource-intensive offline calibration, as shown in Fig.~\ref{fig:teaser}(a)(b). These inaccuracies could be far more exacerbated in real-world applications under computing limits and real-time constraints. Moreover, localization systems are susceptible to long-existing yet still unsolved attacks \cite{noh2019tractor,narain2019security,zeng2018all,shen2020drift,ikram2022perceptual,yoshida2022adversarial}. These attacks allow adversaries to manipulate positions at will, further undermining the reliability of localization systems.
Such prevalent challenges of significant noise and malicious attack starkly contrast with the ideal scenarios considered by earlier works \cite{vadivelu2021learning, yuan2022keypoints,xu2022v2x,lu2023robust}, which primarily focus on minor pose inaccuracies and fail to transcend the no collaboration baseline under large noise, see Fig.~\ref{fig:teaser}(c). 

To eliminate the dependence on potentially unreliable external localization systems, a direct solution is deducing the relative poses of collaborative agents by point cloud registration, a technique extensively utilized in multi-agent collaborative systems\cite{lajoie2023swarm,zhong2024colrio,chang2023hydra,chang2022lamp,zhong2023dcl}. Point cloud registration methods \cite{besl1992method,teaser, ao2021spinnet, huang2021predator} apply nearest-neighbor algorithms to identify correspondences across extensive 3D point sets, followed by robust techniques \cite{ransac,teaser} to calculate the transformation from these putative correspondences. Although these methods prove effective for latency-tolerant applications such as collaborative mapping\cite{lajoie2023swarm}, the real-time transmission of large volumes of 3D data is impractical for bandwidth-limited collaborative perception systems \cite{huang2021disco,hu2022where2comm,chen2019fcooper,xu2023v2v4real}. Therefore, there exists a pronounced gap in creating a system that is free from localization errors while also maintaining communication efficiency for practical applications.

To fill this gap, we propose $\mathtt{CoBEVGlue}$, a self-localized collaborative perception system that is designed for multiple agents to achieve more holistic perception without relying on external localization systems, achieving efficiency with reduced communication costs. $\mathtt{CoBEVGlue}$ follows the pipeline of previous collaborative perception systems \cite{lu2023robust,hu2022where2comm}, and uses its key spatial alignment module $\mathtt{BEVGlue}$ to estimate the relative pose between agents with the objects detected and tracked by each agent.
The core idea behind $\mathtt{BEVGlue}$ is to search for the co-visible objects from the bird’s eye view perceptual data across agents and calculate the relative transformation with these co-visible objects, ensuring a consistent spatial coordinate system for collaboration. $\mathtt{BEVGlue}$ includes three key components: i) object graph modeling, which converts each agent's observations to an object graph with rich information, including object shape, heading, tracking ID and the invariant spatial relationship between objects; ii) temporally consistent maximum subgraph detection, which efficiently harnesses spatial and temporal data within object graphs to detect the largest common subgraph, following strict spatial isomorphism constraint and temporal consistency; and iii) relative pose calculation, which computes the pose relationships between agents using the detected common subgraph, without using time-consuming outlier rejection algorithms.

The proposed $\mathtt{CoBEVGlue}$ system offers three significant advantages: i) it operates independently of external localization devices, showcasing its resilience to noise and malicious attacks; ii) it brings minor communication overhead since $\mathtt{CoBEVGlue}$ only uses object bounding boxes with tracking ID to estimate the relative pose between agents; iii) its core module, $\mathtt{BEVGlue}$ ensures high-quality matching results by keeping strict spatial isomorphism constraint between the detected common subgraph and temporal consistency between matching results across time.

To evaluate the effectiveness of the proposed method, we consider the collaborative 3D object detection task on three datasets: OPV2V \cite{xu2022opv2v}, DAIR-V2X \cite{yu2022dair} and V2V4Real \cite{xu2023v2v4real}, covering both simulation and real-world scenarios. The results show that, the $\mathtt{CoBEVGlue}$ empowered robust collaborative perception system perform comparably to systems relying on precise localization information, and achieves state-of-art detection performance when localization noise and attack exist.

In summary, the main contributions of this work are:
\begin{itemize}
	\item We propose $\mathtt{CoBEVGlue}$, the first self-localized collaborative perception system without relying on external localization devices;
	\item We propose $\mathtt{BEVGlue}$, a novel spatial alignment method that estimates the relative poses between agents through matching co-visible  objects;
	\item We conduct extensive experiments for collaborative LiDAR object detection in simulated and real-world datasets. The results show that i)$\mathtt{CoBEVGlue}$ attains state-of-the-art detection performance in the presence of localization noise. ii)$\mathtt{BEVGlue}$ can seamlessly integrate with a majority of previous methods, enhancing their performance by an average of $57.7\%$.  
\end{itemize}

\section{Related work}
\label{sec:related_work}

\subsection{Collaborative Perception}
As a recent application of multi-agent systems to perception tasks, collaborative perception is emerging \cite{xu2022opv2v, wang2020v2vnet,li2021learning,hu2022where2comm}. To support this area of research, there is a surge of
high-quality datasets, including V2X-Sim \cite{li2022v2x}, OPV2V \cite{xu2022opv2v},
and DAIR-V2X \cite{yu2022dair}. Based on those datasets,
numerous methods have been proposed to handle various
practical issues, such as communication latency \cite{lei2022latency} and communication bandwidth \cite{hu2022where2comm}. In this work, we specifically consider the robustness of localization error and attack.

To gain resistance towards localization noises, previous works consider two main approaches: learning-based and matching-based. Learning-based methods aim to construct robust network architectures to reduce the impact of pose errors. For example, V2VNet (robust) \cite{vadivelu2021learning} designs pose regression, global consistency and
attention aggregation module to correct relative poses and
concentrate on neighbor with less pose error; V2X-ViT \cite{xu2022v2x}
uses multi-scale window attention to capture features in
various ranges. On the other hand, matching-based approaches seek to develop robust frameworks or network architectures. Examples include FPV-RCNN \cite{yuan2022keypoints} and CoAlign \cite{lu2023robust}, which estimate relative poses between agents using an IoU-based matching strategy. However, they can only rectify minor inaccuracies in external localization since these approaches rely on a basically precise initial relative pose. Their performance drops significantly when the noise is large or an attack exists. In contrast, our work considers collaborative perception independent of external localization systems.

\subsection{Point Cloud Registration}
Although the ultimate aim of this paper is to enhance detection capabilities, advancements in point cloud registration methods has inspired us to propose our novel self-localized collaborative perception system. Traditional point cloud registration methods focusing on refining the Iterative Closest Point (ICP) algorithm \cite{besl1992method} and its variants \cite{bouaziz2013sparse,fitzgibbon2003robust,gelfand2005robust,sharp2002icp} have led to improvements in convergence and noise resilience. Recent typical point cloud registration workflows consist of extracting local 3D feature descriptors and conducting registration. For extracting 3D local descriptors, conventional approaches like Fast Point Feature Histograms \cite{rusu2009fast,johnson1999using,rusu2008aligning,tombari2010unique} utilize hand-crafted features. More recent techniques \cite{qi2017pointnet++,qi2017pointnet,ao2021spinnet,choy2019fully,zeng20173dmatch,liu2023density} adopt learning-based methods for this purpose. In terms of registration, traditional approaches often employ nearest-neighbor algorithms for matching and robust optimization for outlier rejection \cite{ransac,teaser}, whereas contemporary deep registration methods \cite{huang2021predator,yew2022regtr,yew2020rpm} leverage self-attention mechanisms \cite{vaswani2017attention} for correspondence determination. SGAligner \cite{sgaligner} pioneers the employment of a pre-constructed 3D scene graph for registration purposes. Nevertheless, similar to preceding strategies, it requires the transmission of dense point clouds and high-dimensional features. These methods are widely applied in latency-tolerant multi-agent systems such as collaborative mapping\cite{lajoie2023swarm} and 3D scene graph generation\cite{chang2023hydra}. However, the collaborative object detection task requires precise relative pose estimation in real time. Unfortunately, the V2X networks struggle to transmit the dense point clouds and feature required by point cloud registration methods in real time. To overcome this limitation, our approach prioritizes object-level registration, representing each object with just eight float numbers. This innovation markedly reduces the bandwidth necessary and computation cost for calculating relative poses among collaborative autonomous vehicles, thus efficiently resolving the transmission dilemma.


\subsection{Maximum Common Subgraph Detection}
The Maximum Common Subgraph (MCS) detection problem, classified as NP-hard, is pivotal in various scientific fields ene\cite{raymond2002maximum,combier2013map,ehrlich2011maximum,gay2014subgraph}, necessitating algorithms that balance precision and computational efficiency. Traditional approaches primarily employ branch-and-bound algorithms \cite{mcgregor1982backtrack, vismara2008finding, mccreesh2017partitioning} and techniques that transform MCS detection into maximum clique problems \cite{raymond2002maximum, mccreesh2016clique}.  Recent advancements \cite{liu2019learning,bai2021glsearch} in machine learning have seen the application of graph neural networks and reinforcement learning to MCS detection, which attempts to learn suitable heuristics for graph matching. Despite their innovations, they are still constrained by the heuristic nature of the search space exploration and are subject to exponential time complexity in the worst-case scenarios. In this work, we model the bounding boxes detected by each agent with a geometric invariant object pose graph and leverage the spatial constraints and temporal consistency to solve the problem efficiently.

\section{CoBEVGlue: Self-Localized Collaborative Perception System}
\label{sec:coop}

\begin{figure*}[h]
\centering
\centering{\includegraphics[width=0.99\linewidth]{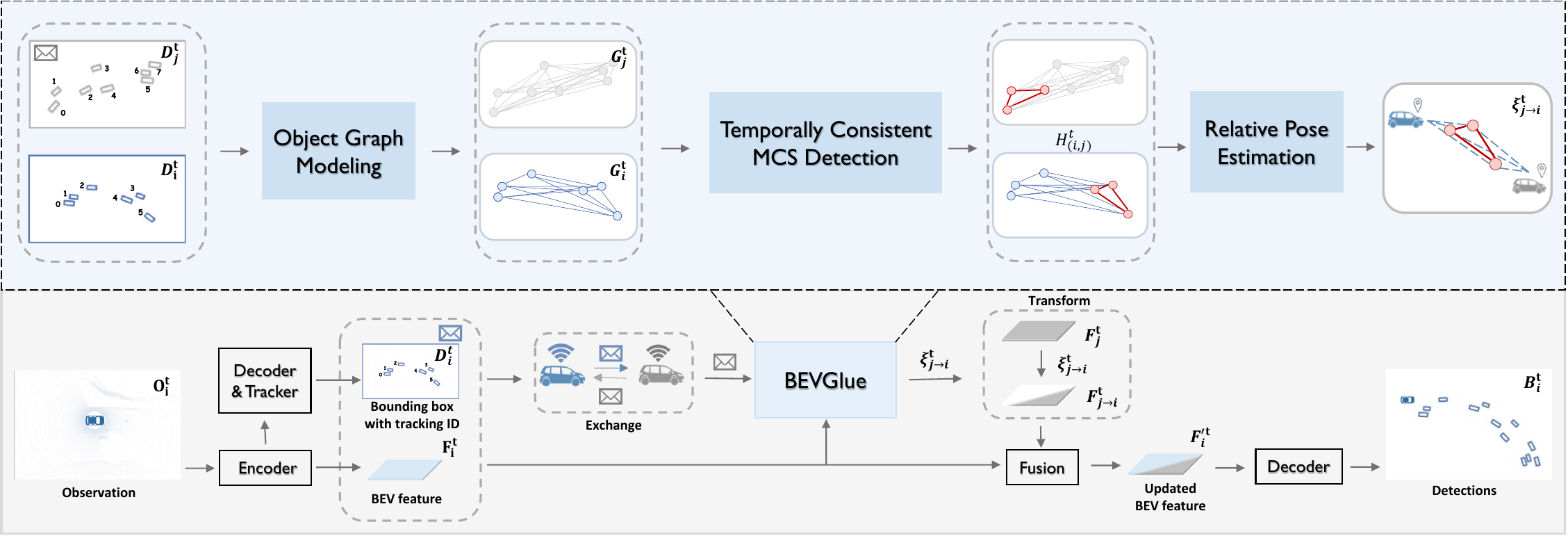}}
\caption{Overview of the proposed self-localized collaborative perception framework. The key module is $\mathtt{BEVGlue}$, which leverages object graphs and temporally consistent MCS detection to achieve spatial alignment.}
\label{fig:overview}
\vspace{-3mm}
\end{figure*}

In this section, we present $\mathtt{CoBEVGlue}$, the first self-localized collaborative perception system that replaces potentially unreliable localization systems with our novel spatial alignment module $\mathtt{BEVGlue}$ to estimate the relative pose between agents. $\mathtt{CoBEVGlue}$ includes a single-agent object detector and tracker, the key spatial alignment module $\mathtt{BEVGlue}$, a multi-agent feature fusion module, and a decoder; see the overview in  Fig.~\ref{fig:overview}.

Mathematically, consider $N$ agents in the scene. For the $i$th agent, let $\mathbf{O}_{i}^t$ be the perceptual observation at time $t$. The proposed $\mathtt{CoBEVGlue}$ works as follows:
\begin{subequations}
    \begin{eqnarray}
    \label{eq:bevglue_encoder}
    \mathbf{F}_i^t, \mathbf{D}_i^t & = & f_{\rm detection\&tracking} \left(\mathbf{O}_{i}^t \right),
    \\
    \label{eq:bevglue_correction}
    \xi_{j \rightarrow i}^t& = & f_{\rm BEVGlue} \left(
    \mathbf{D}_{i}^{t}, \mathbf{D}_{j}^{t} \right),
    \\
    \label{eq:bevglue_transform}
    \mathbf{F}_{j \rightarrow i}^t & = & f_{\rm transform} \left( \mathbf{F}_j^t, \xi_{j \rightarrow i}^t\right),
    \\
    \label{eq:bevglue_fusion}
    \mathbf{F}_i^{'t} & = & f_{\rm fusion} \left( \{ \mathbf{F}_{j \rightarrow i}^t \}_{j=1,2, \cdots,N} \right),
    \\
    \label{eq:bevglue_decoder}
    \mathbf{B}_i^{t}  & = & f_{\rm decoder} \left( \mathbf{F}_i^{'t} \right),
    \end{eqnarray}
\end{subequations}
where $\mathbf{F}_i^t$ is the BEV feature extracted from the $i$th agent's observation, $\mathbf{D}_i^t$ is the detection and tracking outputs without collaboration, $\xi_{j \rightarrow i}^t$ is the estimated relative pose from $i$th agent's perspective to $j$th agent ($\xi_{i \rightarrow i}^t$ is the identity), $\mathbf{F}_{j \rightarrow i}^t$ is the wrapped BEV feature transformed from the $j$th agent's coordinate space to the $i$th agent's coordinate space through affine transformation, $\mathbf{F}_{i}^{'t}$ is the aggregated feature of the $i$th agent after fusing other agents' messages, and $\mathbf{B}_i^{t}$ is the detection outputs after collaboration.

Step~\eqref{eq:bevglue_encoder} employs the PointPillar framework \cite{lang2019pointpillars}, a lightweight 3D object detection system, in conjunction with a SORT \cite{bewley2016simple}-inspired tracker, to extract the BEV feature $\mathbf{F}_i^t$ from the observation $\mathbf{O}_i^t$ of the $i$th agent. This step also generates the detected bounding boxes accompanied by their tracking IDs $\mathbf{D}_i^t$. The cornerstone of the process, Step~\eqref{eq:bevglue_correction}, leverages our innovative $\mathtt{BEVGlue}$ module to identify co-visible objects and compute the relative pose $\xi_{j \rightarrow i}^t$, drawing upon the detection and tracking results from multiple agents; see details in the Section \ref{sec:bevglue}. Subsequently, Step~\eqref{eq:bevglue_transform} aligns the features from other agents with the ego agent's pose using the estimated relative poses. Step~\eqref{eq:bevglue_fusion} applies a multi-scale max fusion to refine the feature map, denoting $\mathbf{F}_{i \rightarrow i}^t$ as equivalent to $\mathbf{F}_i^t$. The final phase, Step~\eqref{eq:bevglue_decoder}, uses fused features to obtain final detection results. $\mathtt{CoBEVGlue}$ can be applied to multi-agent collaboration: Steps~\eqref{eq:bevglue_correction} and ~\eqref{eq:bevglue_transform} are executed between the ego agent and each collaborator independently. Subsequently, in Step~\eqref{eq:bevglue_fusion}, the ego agent integrates features transformed from all $N$ agents.

Note that accurate relative pose estimation, $\xi_{j \rightarrow i}^t$, is essential for the success of collaborative perception systems. Inaccuracies in pose information can critically undermine subsequent processes such as feature transformation, fusion, and collaborative detection. Conventionally, as delineated in Step~\eqref{eq:bevglue_correction}, precise pose information necessitates each agent to utilize an external localization system to acquire its global position and calculate the relative transformations among collaborators. This dependency on external localization is fraught with challenges, including susceptibility to noise interference and potential security breaches through malicious attacks. Our innovative spatial alignment module, $\mathtt{BEVGlue}$, is designed to solve these issues by leveraging perceptual data to ensure accurate relative pose estimation, thereby enhancing the resilience and effectiveness of collaborative perception. We elaborate this key module in Sec. \ref{sec:bevglue}.

\vspace{-2mm}

\section{BEVGlue: Spatial Alignment Module}

\vspace{-1mm}

\label{sec:bevglue}
To estimate the relative pose $\xi_{j \rightarrow i}^t$ between agents, the main idea of $\mathtt{BEVGlue}$ is to identify the co-visible objects and subsequently calculate the transformation based on these co-visible objects. To excavate this internal correspondence among agents, $\mathtt{BEVGlue}$ presents three modules: (i) object graph modeling, (ii) temporally consistent maximum common subgraph detection, and (iii) relative pose calculation.

\subsection{Object Graph Modeling}

Object graph modeling is designed to represent the detection and tracking outcomes of each agent from Step~\eqref{eq:bevglue_encoder} as an object graph, with each node corresponding to an object and each edge describing the spatial relationship between objects. This method of modeling node and edge attributes discovers the temporal information of each object and the invariant geometric pattern between objects, which are valuable information for the subsequent common subgraph searching procedure.

\begin{figure}[h]
\centering
\centering{\includegraphics[width=0.99\linewidth]{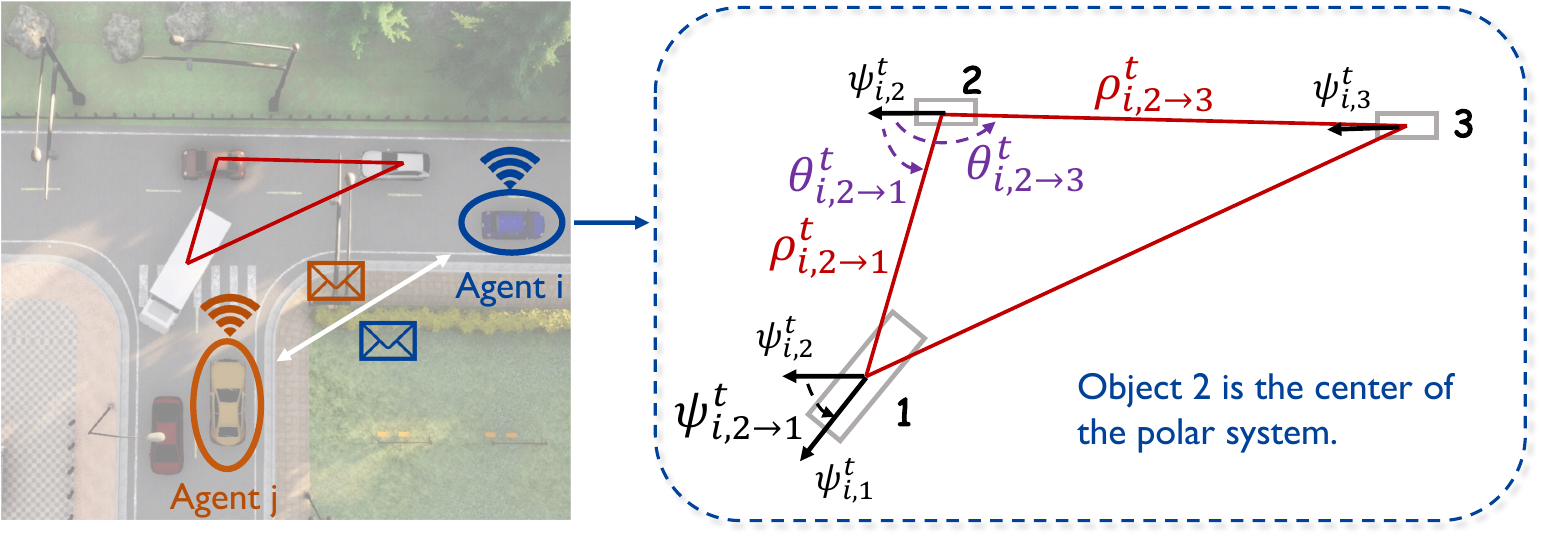}}
\caption{Illustration of the proposed object graph modeling. The \textcolor{blue}{Ego agent} and its \textcolor{orange}{collaborator} is collaborating at a T-junction. Three co-visible object are connected by \textcolor{red}{red} line, representing their spatial relationship. The right portion visualizes the meaning of variables in the edge features. }
\label{fig:bevglue}
\vspace{-3mm}
\end{figure}

Consider the $i$th agent is tracking $M_i$ objects in a scenario. 
Let $\mathbf{D}^t_i = [\mathbf{b}^t_{i,1}, \mathbf{b}^t_{i,2}, ..., \mathbf{b}^t_{i,M_i}] \in \mathbb{R}^{M_i \times 6}$ be the detection and tracking result. The $m$th bounding box with tracking ID is $\mathbf{b}^t_{i,m} = [x^t_{i,m}, y^t_{i,m}, l^t_{i,m}, w^t_{i,m}, \psi^t_{i,m}, \tau^t_{i,m}]$, encompassing the 2D center position, length, width, yaw angle, and tracking ID. We formulate $\mathbf{D}^t_i$ into a fully connected object graph $\mathcal{G}^t_i(\mathcal{V}^t_i, \mathcal{E}^t_i)$ for agent $i$, where $\mathcal{V}^t_i, \mathcal{E}^t_i$ are the sets of nodes and edges, respectively. To be noted that each agent has its object graph. The $m$th node feature $\mathbf{v}^t_{i,m}$ is $[l^t_{i,m}, w^t_{i,m}, \tau^t_{i,m}]$, and the edge feature is $\mathbf{e}^t_{i,mn} =[\rho^t_{i,m \rightarrow n}, \theta^t_{i,m \rightarrow n}, \psi^t_{i,m \rightarrow n}]$ 
$\rho^t_{i,m \rightarrow n}, \theta^t_{i,m \rightarrow n} \ \text{and}\ \psi^t_{i,m \rightarrow n}$ are defined within a polar coordinate system which sets the heading of the $m$th node as the reference direction and its position as the origin (pole). The $\rho^t_{i,m \rightarrow n}$ and $\theta^t_{i,m \rightarrow n}$ denote the radial distance and polar angle of node $n$ respectively and $\psi^t_{i,m \rightarrow n}$ is the intersection angle between the heading of node $n$ and node $m$.
Given that the pole and the reference direction possess clear and singular definitions in the physical world, consistency in edge feature computation across different object graphs is achievable. Specifically, if the detection results of nodes $m$ and $n$ are accurate for both the $i$th and $j$th agents when calculating the edge feature $\mathbf{e}^t_{j,mn}$ on agent graph $\mathcal{G}_j^t$ based on agent $j$, it will be identical to $\mathbf{e}^t_{i,mn}$.
Also see the Fig.~\ref{fig:bevglue}.

The object graph presents an innovative approach to model the observation of each agent: i) the node attribute encompasses temporal tracking data, which helps keep the matching consistency across time; ii) the edge feature is consistent across object graphs derived from different agents' perspectives, signifying that rotations and translations applied to $\mathbf{D}^t_i$ do not alter the value of $\mathbf{e}^t_{i,mn}$. It implies that when two objects are simultaneously observed by different agents, the edge attribute remains consistent, regardless of the varying perspectives.

\subsection{Temporally Consistent Maximum Common Subgraph Detection}
Upon completing the object graph modeling phase, the task shifts to efficiently detecting the largest common subgraph that strictly satisfies graph isomorphism between two graphs. This common subgraph is indicative of the co-visible objects across agents and is subsequently used in calculating the relative transformation. We leverage two pieces of information for the search of a common subgraph: i) the spatial relationship: the geometric pattern of co-visible objects across different agents is isomorphism; ii) the temporal relationship: the co-visible objects are consistent across time. By leveraging the spatial relationship, we can ensure there are no outliers in the matched nodes. By leveraging the temporal relationship, we can ensure the temporal consistency of matching results across time.

Consider a pair of modeled object pose graphs $\mathcal{G}_i^t = \{\mathcal{V}_i^t,\mathcal{E}_i^t\}$ with $M$ nodes and $\mathcal{G}_j^t = \{\mathcal{V}_j^t,\mathcal{E}_j^t\}$ with $N$ nodes at time $t$, finding the maximum common subgraph $\mathcal{H}_{(i,j)}^{t}$ can be formulated as
\begin{equation}
    \mathcal{H}_{(i,j)}^{t} = f_{\text{MCS}}\left(\mathcal{G}_i^t, \mathcal{G}_j^t, \mathcal{H}_{(i,j)}^{t-1}\right).
\end{equation}

To realize $f_{\text{MCS}}\left(\cdot\right)$, the procedure is divided into three primary steps:

\textbf{i). Candidate initialization}. 
This step is designed to generate a set of candidate common subgraphs, represented as a list $\mathcal{S}^t = [s_0^t, s_1^t, ..., s_K^t]$, where $K$ denotes the number of these candidates. At the initial timestep($t=0$), we explore all $M \times N$ potential node pairs. For each pair, we assess whether their node affinity surpasses a predefined threshold $\gamma_{\mathcal{V}}$. 
The node affinity for any given node pair $(m,n)$ at timestep $t$ is determined by the equation $\kappa_{mn}^{\mathcal{V}^t,(i,j)} = \phi_{\mathcal{V}}(\mathbf{v}^t_{i,m},\mathbf{v}^t_{j,n})$ where the affinity function $\phi_{\mathcal{V}}\left(\cdot\right)$ measure the similarity between node $m$ in agent graph $\mathcal{G}_i^t$ and node $n$ in agent graph $\mathcal{G}_j^t$. Pairs that meet this criterion are subsequently incorporated into $\mathcal{S}^t$ as candidates for common subgraphs.
At subsequent timesteps($t\geq 1$), the node pairs from previous common subgraph $\mathcal{H}_{(i,j)}^{t-1}$ are considered as candidates for the current common subgraphs. The temporal correspondence between $\mathcal{S}^t$ and $\mathcal{H}_{(i,j)}^{t-1}$ is established by the tracking IDs.

\textbf{ii). Subgraph expanding}. For each candidate $s_k^t$, we consider the node pair within it as $(v^t_{i,p},v^t_{j,q})$. We then identify all potential matching node pairs $(v^t_{i,m},v^t_{j,n})$ using node affinity $\kappa_{mn}^{\mathcal{V}^t,(i,j)} = \phi_{\mathcal{V}}(\mathbf{v}^t_{i,m},\mathbf{v}^t_{j,n})$ and edge affinity $\kappa_{(p,m),(q,n)}^{\mathcal{E}^t,(i,j)} = \phi_{\mathcal{E}}(\mathbf{e}^t_{i,pm},\mathbf{e}^t_{j,qn})$. The affinity functions for node is the same as i) and the affinity functions for edges $\phi_{\mathcal{E}}\left(\cdot\right)$ measure the similarity between edge pairs, considering the difference of relative position and heading difference; see the details in Appendix. A pair $(v^t_{i,m},v^t_{j,n})$ will be added to $s_k^t$ if $\kappa_{mn}^{\mathcal{V}^t,(i,j)} > \gamma_{\mathcal{V}}$ and $\kappa_{(p,m),(q,n)}^{\mathcal{E}^t,(i,j)} > \gamma_{\mathcal{E}}$ where $\gamma_{\mathcal{V}}$ and $\gamma_{\mathcal{E}}$ are predefined thresholds.

\textbf{iii). Maximum common subgraph selection}. After ii), our algorithm needs to determine the most suitable common subgraph from all candidates in $\mathcal{S}^t$. The initial criterion is to select the common subgraphs with the highest node count. If multiple candidates share this characteristic, we then compute a confidence score for those candidates.
The confidence score $c_k^t$ of the candidate $s_k^t=\{(v^t_{i,p_1},v^t_{j,q_1}),(v^t_{i,p_2},v^t_{j,q_2}),...,(v^t_{i,p_C},v^t_{j,q_C})\}$ is calculated using Eq.~\ref{eq:ckt}.
\begin{equation}
    c_k^t = \sum_{\alpha=1}^C\kappa_{p_\alpha q_\alpha}^{\mathcal{V}^t,(i,j)} + \sum_{\alpha=1}^C\sum_{\beta=1}^C\kappa_{(p_\alpha,p_\beta),(q_\alpha,q_\beta)}^{\mathcal{E}^t,(i,j)},
    \label{eq:ckt}
\end{equation}
where $C$ is the node count of $s_k^t$. Among the candidates that have the largest number of nodes, the one with the highest confidence score is selected as the final common subgraph.

The proposed MCS detection algorithm brings two benefits: i) the subgraph expanding step maintains strict spatial constraints between matched objects, saving the time for outlier detection in the following transformation calculation step. ii) the temporal tracking information promotes the consistency of detected MCS across time, improving the robustness of the proposed algorithm; see the algorithm diagram in appendix.

\subsection{Relative Pose Calculation}
Given the common subgraph $\mathcal{G}_{s,(i,j)}^{t}$, we calculate the relative pose $\xi_{j \rightarrow i}^t$ by considering each matched node as a point and estimate the rigid transformation between two point sets.
Let $\mathcal{P} = [v^t_{i,p_1},v^t_{i,p_2},...v^t_{i,p_C}], \mathcal{Q} = [v^t_{j,q_1},v^t_{j,q_2},...v^t_{j,q_C}]$ be two matching node sets and $\mathbf{P} = [\mathbf{p}_1, \mathbf{p}_2,...,\mathbf{p}_C]$ and $\mathbf{Q} = [\mathbf{q}_1, \mathbf{q}_2,...,\mathbf{q}_C]$ be the corresponding 2D position of the nodes in their corresponding BEV coordinate system. The transformation between two coordinate systems $\mathbf{R}\in SO(2), t \in \mathbb{R}^2$ can be calculated by solving the following Procrustes problem\cite{gower2004procrustes}:
\begin{equation}
\label{equ:svd}
\mathbf{R}^*, \mathbf{t}^*=\mathop{\arg\min}_{\mathbf{R},\mathbf{t}} \sum_{i=1}^C\left\|\mathbf{R}\mathbf{p}_i+\mathbf{t}-\mathbf{q}_i\right\|^2.
\end{equation}

The optimal solution for the transformation can be efficiently derived using Singular Value Decomposition (SVD) in scenarios where there are no erroneous matching pairs. However, in cases where erroneous correspondences are present, the least variance nature of optimizing Eq.~\ref{equ:svd} necessitates the use of other more time-intensive methods capable of outlier elimination, such as RANSAC \cite{ransac}. Thanks to the accuracy of our temporally consistent maximum common subgraph detection, we can employ the faster SVD method for computing the transformation. 
The final relative pose $\xi_{j \rightarrow i}^t$ can be obtained from $\mathbf{R}, t$.

\subsection{Discussions}
\textbf{Advantages.} 
$\mathtt{BEVGlue}$ has following distinct advantages.
Comparing to point cloud registration methods\cite{sharp2002icp,teaser,zeng20173dmatch}:
\begin{itemize}
    \item It only requires the transmission of bounding boxes and tracking IDs, while point cloud registration methods require the transmission of massive points with their feature vector, costing too much communication bandwidth;
    \item It ensures high-quality matching with no outliers by checking the spatial relationship between matched nodes, while point cloud registration methods only match by comparing local point features; 
\end{itemize}
Comparing to alignment modules in previous collaborative perception system methods~\cite{vadivelu2021learning, yuan2022keypoints,xu2022v2x,lu2023robust} :
\begin{itemize}
    \item It is capable of building a consistent coordinate system without using any localization results, while previous methods can only rectify minor inaccuracies in external localization systems since they rely on a basically precise initial relative pose.
    \item It incorporates tracking results to enhance the temporal consistency of matching, while previous methods ignore the temporal information.
\end{itemize}

\textbf{Prerequisites.} 
There are two assumptions for $\mathtt{BEVGlue}$ to function well. i) Collaboration is initiated only when agents are in close proximity, ensuring a common field of view. ii) Several objects exist in the common field of view and are perceived by both agents. 
These assumptions are realistically met in a collaborative perception context for two primary reasons. Firstly, the pivotal wireless connections that facilitate collaboration are inherently range-bound, thus naturally restricting collaborative interactions to agents in close proximity. Secondly, typical traffic scenarios inherently offer an environment rich with various objects and BEV alignment can be successfully achieved with the presence of only two co-visible devices. Real-world data from collaborative perception datasets corroborate this: 91.3\% of the samples in DAIR-V2X\cite{yu2022dair} and 94.2\% of the samples in V2V4Real\cite{xu2023v2v4real} satisfy these criteria.

\section{Experimental Results}
\label{sec:experiment}

\begin{table*}[!t]
\centering
\caption{Detection performance with localization noises following Gaussian distribution in the testing phases.} \vspace{-1mm}
\resizebox{\textwidth}{!}{%
\begin{tabular}{clcccccccccccc}
\hline
\multicolumn{2}{c|}{Dataset}                                                                                                                           & \multicolumn{4}{c|}{OPV2V}                                                               & \multicolumn{4}{c|}{DAIR-V2X}                                                            & \multicolumn{4}{c}{V2V4Real}                                        \\ \hline
\multicolumn{2}{c|}{Method/Metric}                                                                                                                     & \multicolumn{12}{c}{AP@0.5 $\uparrow$}                                                                                                                                                                                                                    \\ \hline
\multicolumn{2}{c|}{Noise Level $\sigma_t/\sigma_r(m/^{\circ})$}                                                                                       & 0.0/0.0                               & 0.5/0.5 & 1.5/1.5 & \multicolumn{1}{c|}{2.5/2.5} & 0.0/0.0                               & 0.5/0.5 & 1.5/1.5 & \multicolumn{1}{c|}{2.5/2.5} & 0.0/0.0                               & 0.5/0.5 & 1.5/1.5 & 2.5/2.5 \\ \hline
\multicolumn{1}{c|}{w/o collaboration}                                                               & \multicolumn{1}{l|}{No Collaboration}           & \multicolumn{4}{c|}{0.786}                                                               & \multicolumn{4}{c|}{0.645}                                                               & \multicolumn{4}{c}{0.447}                                           \\ \hline
\multicolumn{1}{c|}{}                                                                                & \multicolumn{1}{l|}{F-Cooper\cite{chen2019fcooper}\textcolor{blue}{SEC'19}}                   & {\color[HTML]{656565} 0.834}          & 0.638   & 0.458   & \multicolumn{1}{c|}{0.399}   & {\color[HTML]{656565} 0.737}          & 0.697   & 0.660   & \multicolumn{1}{c|}{0.636}   & {\color[HTML]{656565} 0.693}          & 0.481   & 0.330   & 0.309   \\
\multicolumn{1}{c|}{}                                                                                & \multicolumn{1}{l|}{V2VNet\cite{wang2020v2vnet}\textcolor{blue}{ECCV'20}}                     & {\color[HTML]{656565} 0.936}          & 0.861   & 0.724   & \multicolumn{1}{c|}{0.691}   & {\color[HTML]{656565} 0.665}          & 0.610   & 0.551   & \multicolumn{1}{c|}{0.526}   & {\color[HTML]{656565} 0.580}          & 0.441   & 0.338   & 0.312   \\
\multicolumn{1}{c|}{}                                                                                & \multicolumn{1}{l|}{DiscoNet\cite{li2021learning}\textcolor{blue}{NeurIPS'21}}                   & {\color[HTML]{656565} 0.916}          & 0.874   & 0.788   & \multicolumn{1}{c|}{0.753}   & {\color[HTML]{656565} 0.737}          & 0.704   & 0.674   & \multicolumn{1}{c|}{0.666}   & {\color[HTML]{656565} \textbf{0.736}} & 0.527   & 0.411   & 0.378   \\
\multicolumn{1}{c|}{\multirow{-4}{*}{\begin{tabular}[c]{@{}c@{}}w/o\\ robust\\ design\end{tabular}}} & \multicolumn{1}{l|}{Where2comm\cite{hu2022where2comm}\textcolor{blue}{NeurIPS'22}}                 & {\color[HTML]{656565} 0.944}          & 0.721   & 0.500   & \multicolumn{1}{c|}{0.505}   & {\color[HTML]{656565} \textbf{0.752}} & 0.637   & 0.580   & \multicolumn{1}{c|}{0.570}   & {\color[HTML]{656565} 0.704}          & 0.505   & 0.384   & 0.364   \\ \hline
\multicolumn{1}{c|}{}                                                                                & \multicolumn{1}{l|}{FPV-RCNN\cite{yuan2022keypoints}\textcolor{blue}{RAL'22}}                   & {\color[HTML]{656565} 0.858}          & 0.476   & 0.236   & \multicolumn{1}{c|}{0.225}   & {\color[HTML]{656565} 0.626}          & 0.512   & 0.427   & \multicolumn{1}{c|}{0.422}   & {\color[HTML]{656565} 0.701}          & 0.387   & 0.244   & 0.237   \\
\multicolumn{1}{c|}{}                                                                                & \multicolumn{1}{l|}{V2VNet$_{\text{robust}}$\cite{vadivelu2021learning}\textcolor{blue}{CoRL'20}} & {\color[HTML]{656565} 0.942}          & 0.919   & 0.865   & \multicolumn{1}{c|}{0.831}   & {\color[HTML]{656565} 0.661}          & 0.639   & 0.614   & \multicolumn{1}{c|}{0.594}   & {\color[HTML]{656565} 0.550}          & 0.525   & 0.467   & 0.435   \\
\multicolumn{1}{c|}{}                                                                                & \multicolumn{1}{l|}{V2X-ViT\cite{xu2022v2x}\textcolor{blue}{ECCV'22}}                    & {\color[HTML]{656565} 0.946}          & 0.925   & 0.796   & \multicolumn{1}{c|}{0.632}   & {\color[HTML]{656565} 0.705}          & 0.682   & 0.647   & \multicolumn{1}{c|}{0.632}   & {\color[HTML]{656565} 0.680}          & 0.673   & 0.450   & 0.422   \\
\multicolumn{1}{c|}{\multirow{-4}{*}{\begin{tabular}[c]{@{}c@{}}w/\\ robust\\ design\end{tabular}}}  & \multicolumn{1}{l|}{CoAlign\cite{lu2023robust}\textcolor{blue}{ICRA'23}}                    & {\color[HTML]{656565} \textbf{0.966}} & 0.950   & 0.863   & \multicolumn{1}{c|}{0.824}   & {\color[HTML]{656565} 0.746}          & 0.712   & 0.665   & \multicolumn{1}{c|}{0.647}   & {\color[HTML]{656565} 0.709}          & 0.613   & 0.435   & 0.387   \\ \hline
\multicolumn{1}{c|}{Self-Localized}                                                                  & \multicolumn{1}{l|}{CoBEVGlue}                  & \multicolumn{4}{c|}{\textbf{0.958}}                                                      & \multicolumn{4}{c|}{\textbf{0.740}}                                                      & \multicolumn{4}{c}{\textbf{0.702}}                                  \\ \hline
\multicolumn{14}{l}{}                                                                                                                                                                                                                                                                                                                                                                                              \\ \hline
\multicolumn{2}{c|}{Method/Metric}                                                                                                                     & \multicolumn{12}{c}{AP@0.7 $\uparrow$}                                                                                                                                                                                                                    \\ \hline
\multicolumn{2}{c|}{Noise Level $\sigma_t/\sigma_r(m/^{\circ})$}                                                                                       & 0.0/0.0                               & 0.5/0.5 & 1.5/1.5 & \multicolumn{1}{c|}{2.5/2.5} & 0.0/0.0                               & 0.5/0.5 & 1.5/1.5 & \multicolumn{1}{c|}{2.5/2.5} & 0.0/0.0                               & 0.5/0.5 & 1.5/1.5 & 2.5/2.5 \\ \hline
\multicolumn{1}{c|}{w/o collaboration}                                                               & \multicolumn{1}{l|}{No Collaboration}           & \multicolumn{4}{c|}{0.690}                                                               & \multicolumn{4}{c|}{0.526}                                                               & \multicolumn{4}{c}{0.261}                                           \\ \hline
\multicolumn{1}{c|}{}                                                                                & \multicolumn{1}{l|}{F-Cooper\cite{chen2019fcooper}\textcolor{blue}{SEC'19}}                   & {\color[HTML]{656565} 0.603}          & 0.388   & 0.328   & \multicolumn{1}{c|}{0.298}   & {\color[HTML]{656565} 0.560}          & 0.542   & 0.516   & \multicolumn{1}{c|}{0.487}   & {\color[HTML]{656565} 0.432}          & 0.212   & 0.179   & 0.174   \\
\multicolumn{1}{c|}{}                                                                                & \multicolumn{1}{l|}{V2VNet\cite{wang2020v2vnet}\textcolor{blue}{ECCV'20}}                     & {\color[HTML]{656565} 0.740}          & 0.534   & 0.384   & \multicolumn{1}{c|}{0.315}   & {\color[HTML]{656565} 0.402}          & 0.362   & 0.320   & \multicolumn{1}{c|}{0.316}   & {\color[HTML]{656565} 0.250}          & 0.163   & 0.135   & 0.130   \\
\multicolumn{1}{c|}{}                                                                                & \multicolumn{1}{l|}{DiscoNet\cite{li2021learning}\textcolor{blue}{NeurIPS'21}}                   & {\color[HTML]{656565} 0.791}          & 0.741   & 0.684   & \multicolumn{1}{c|}{0.655}   & {\color[HTML]{656565} 0.584}          & 0.568   & 0.561   & \multicolumn{1}{c|}{0.557}   & {\color[HTML]{656565} 0.466}          & 0.296   & 0.271   & 0.357   \\
\multicolumn{1}{c|}{\multirow{-4}{*}{\begin{tabular}[c]{@{}c@{}}w/o\\ robust\\ design\end{tabular}}} & \multicolumn{1}{l|}{Where2comm\cite{hu2022where2comm}\textcolor{blue}{NeurIPS'22}}                 & {\color[HTML]{656565} 0.855}          & 0.469   & 0.355   & \multicolumn{1}{c|}{0.286}   & {\color[HTML]{656565} 0.588}          & 0.473   & 0.454   & \multicolumn{1}{c|}{0.451}   & {\color[HTML]{656565} 0.469}          & 0.263   & 0.226   & 0.220   \\ \hline
\multicolumn{1}{c|}{}                                                                                & \multicolumn{1}{l|}{FPV-RCNN\cite{yuan2022keypoints}\textcolor{blue}{RAL'22}}                   & {\color[HTML]{656565} 0.840}          & 0.214   & 0.173   & \multicolumn{1}{c|}{0.189}   & {\color[HTML]{656565} 0.409}          & 0.319   & 0.325   & \multicolumn{1}{c|}{0.340}   & {\color[HTML]{656565} \textbf{0.479}} & 0.153   & 0.156   & 0.165   \\
\multicolumn{1}{c|}{}                                                                                & \multicolumn{1}{l|}{V2VNet$_{\text{robust}}$\cite{vadivelu2021learning}\textcolor{blue}{CoRL'20}} & {\color[HTML]{656565} 0.854}          & 0.826   & 0.773   & \multicolumn{1}{c|}{0.742}   & {\color[HTML]{656565} 0.486}          & 0.472   & 0.447   & \multicolumn{1}{c|}{0.449}   & {\color[HTML]{656565} 0.309}          & 0.296   & 0.279   & 0.272   \\
\multicolumn{1}{c|}{}                                                                                & \multicolumn{1}{l|}{V2X-ViT\cite{xu2022v2x}\textcolor{blue}{ECCV'22}}                    & {\color[HTML]{656565} 0.856}          & 0.834   & 0.721   & \multicolumn{1}{c|}{0.502}   & {\color[HTML]{656565} 0.531}          & 0.523   & 0.510   & \multicolumn{1}{c|}{0.502}   & {\color[HTML]{656565} 0.391}          & 0.305   & 0.272   & 0.262   \\
\multicolumn{1}{c|}{\multirow{-4}{*}{\begin{tabular}[c]{@{}c@{}}w/\\ robust\\ design\end{tabular}}}  & \multicolumn{1}{l|}{CoAlign\cite{lu2023robust}\textcolor{blue}{ICRA'23}}                    & {\color[HTML]{656565} \textbf{0.912}} & 0.878   & 0.771   & \multicolumn{1}{c|}{0.732}   & {\color[HTML]{656565} \textbf{0.604}} & 0.575   & 0.558   & \multicolumn{1}{c|}{0.548}   & {\color[HTML]{656565} 0.417}          & 0.336   & 0.261   & 0.239   \\ \hline
\multicolumn{1}{c|}{Self-Localized}                                                                  & \multicolumn{1}{l|}{CoBEVGlue}                  & \multicolumn{4}{c|}{\textbf{0.909}}                                                      & \multicolumn{4}{c|}{\textbf{0.582}}                                                      & \multicolumn{4}{c}{\textbf{0.431}}                                  \\ \hline
\end{tabular}
\vspace{-3mm}
}
\label{table:noise}
\end{table*}

\begin{table*}[t]
\centering
\caption{With $\mathtt{BEVGlue}$, a majority of representative collaborative perception systems significantly improve their robustness to localization attack, bringing negligible communication overhead. $\delta b$ represents the extra communication overhead required by $\mathtt{BEVGlue}$. } \vspace{-1mm}
\resizebox{\textwidth}{!}{

\begin{tabular}{lllllll}
\hline
Dataset                                             & \multicolumn{3}{c}{OPV2V}                               & \multicolumn{3}{c}{DAIR-V2X}          \\ \hline
\multicolumn{1}{l|}{\multirow{2}{*}{Method/Metric}} & \multicolumn{1}{c}{AP@0.5 $\uparrow$}     & \multicolumn{1}{c}{AP@0.7 $\uparrow$}     & \multicolumn{1}{c}{$\delta b \downarrow$}        & \multicolumn{1}{c}{AP@0.5 $\uparrow$}     & \multicolumn{1}{c}{AP@0.7  $\uparrow$}    & \multicolumn{1}{c}{$\delta b \downarrow$ } \\
\multicolumn{1}{l|}{}                               & \multicolumn{6}{c}{ \textbf{without / with BEVGlue} }                                               \\ \hline
\multicolumn{1}{l|}{F-Cooper\cite{chen2019fcooper}\textcolor{blue}{SEC'19}} & 0.307/0.841 \color{red}{$\uparrow$174\%} & 0.224/0.605 \color{red}{$\uparrow$170\%}   & \multicolumn{1}{l|}{$\uparrow$ 0.000360\%} & 0.563/0.699 \color{red}{$\uparrow$24\%} & 0.410/0.540 \color{red}{$\uparrow$32\%}  & $\uparrow$ 0.000300\%          \\
\multicolumn{1}{l|}{V2VNet\cite{wang2020v2vnet}\textcolor{blue}{ECCV'20}}  & 0.636/0.929 \color{red}{$\uparrow$46\%} & 0.375/0.731 \color{red}{$\uparrow$95\%} & \multicolumn{1}{l|}{$\uparrow$ 0.00144\%}  & 0.393/0.616 \color{red}{$\uparrow$57\%}  & 0.247/0.374 \color{red}{$\uparrow$51\%}  & $\uparrow$ 0.00168\%         \\
\multicolumn{1}{l|}{DiscoNet\cite{li2021learning}\textcolor{blue}{NeurIPS'21}} & 0.671/0.917 \color{red}{$\uparrow$37\%} & 0.654/0.789 \color{red}{$\uparrow$21\%} & \multicolumn{1}{l|}{$\uparrow$ 0.000360\%} & 0.635/0.706 \color{red}{$\uparrow$11\%} & 0.540/0.569 \color{red}{$\uparrow$5.4\%}  & $\uparrow$ 0.000420\%         \\
\multicolumn{1}{l|}{Where2comm\cite{hu2022where2comm}\textcolor{blue}{NeurIPS'22}} & 0.272/0.937 \color{red}{$\uparrow$244\%} & 0.203/0.826 \color{red}{$\uparrow$307\%}  & \multicolumn{1}{l|}{$\uparrow$ 0.000363\%} & 0.493/0.669 \color{red}{$\uparrow$36\%} & 0.398/0.508 \color{red}{$\uparrow$28\%} &  $\uparrow$ 0.000418\%         \\
\multicolumn{1}{l|}{V2VNet$_{\text{robust}}$\cite{vadivelu2021learning}\textcolor{blue}{CoRL'20}} & 0.790/0.927 \color{red}{$\uparrow$17\%} & 0.708/0.837 \color{red}{$\uparrow$18\%}     & \multicolumn{1}{l|}{$\uparrow$ 0.00144\%} & 0.513/0.684 \color{red}{$\uparrow$33\%} & 0.401/0.525 \color{red}{$\uparrow$31\%} & $\uparrow$ 0.00168\%       \\
\multicolumn{1}{l|}{V2X-ViT\cite{xu2022v2x}\textcolor{blue}{ECCV'22}} & 0.696/0.942 \color{red}{$\uparrow$35\%} & 0.638/0.852 \color{red}{$\uparrow$34\%}    & \multicolumn{1}{l|}{$\uparrow$ 0.00150\%} & 0.581/0.684 \color{red}{$\uparrow$18\%} & 0.474/0.525 \color{red}{$\uparrow$11\%}  & $\uparrow$ 0.00172\%      \\
\multicolumn{1}{l|}{CoAlign\cite{lu2023robust}\textcolor{blue}{ICRA'23}} & 0.791/0.962 \color{red}{$\uparrow$22\%} & 0.699/0.907 \color{red}{$\uparrow$30\%}  & \multicolumn{1}{l|}{$\uparrow$ 0.000829\%} & 0.601/0.713 \color{red}{$\uparrow$19\%} & 0.522/0.578 \color{red}{$\uparrow$11\%}  & $\uparrow$ 0.000959\%  \\ \hline
\end{tabular}
\vspace{-3mm}
}
\label{tab:attack}
\end{table*}

\begin{wrapfigure}{R}{0.5\textwidth}
    \centering
    \includegraphics[width=0.5\textwidth]{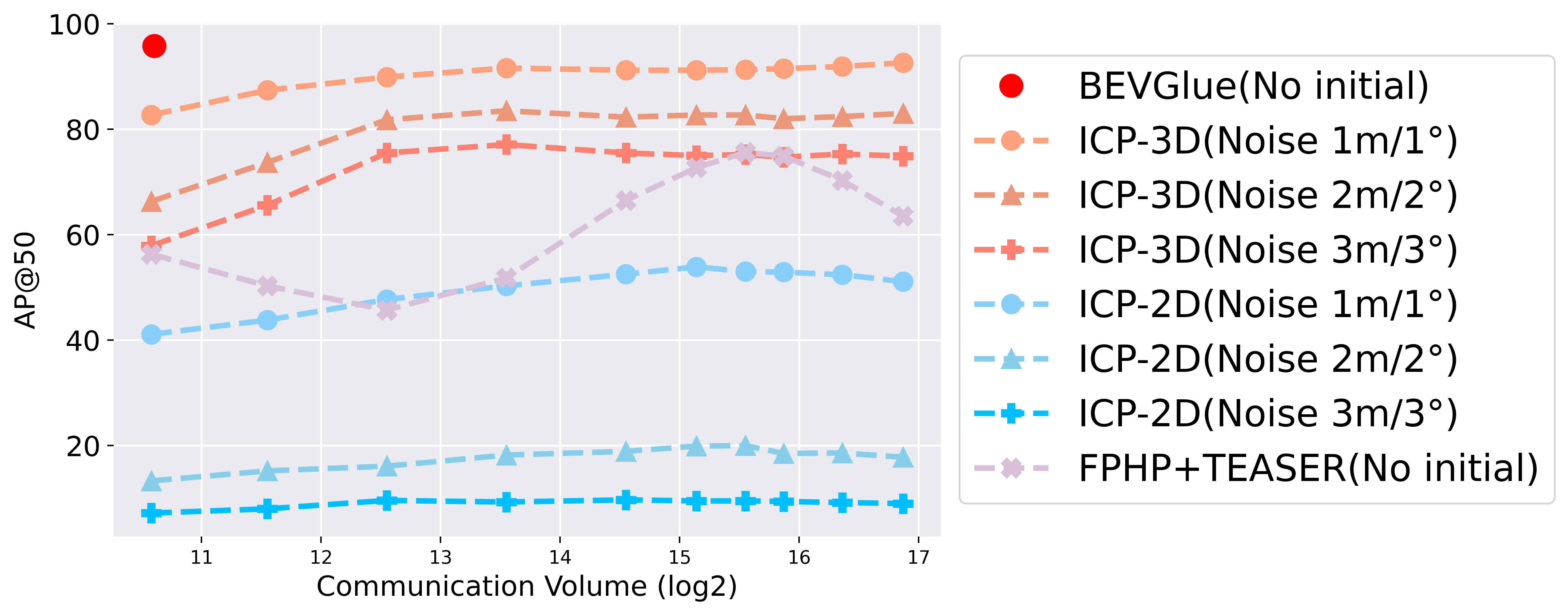}
    \caption{Comparison of collaborative perception with our $\mathtt{BEVGlue}$  and using point cloud registration. $\mathtt{BEVGlue}$ shows significant advantages over baseline methods in both detection and communication efficiency, operating without the need for an initial pose. Initial poses provided to ICP are varied under different Gaussian noise. We sample feature points to conform within the predefined communication volume constraint.}
    \label{fig:icp}
       \vspace{-8mm}
\end{wrapfigure}


\begin{figure*}[t]
\begin{subfigure}{.32\textwidth}
  \centering
  \includegraphics[width=\linewidth]{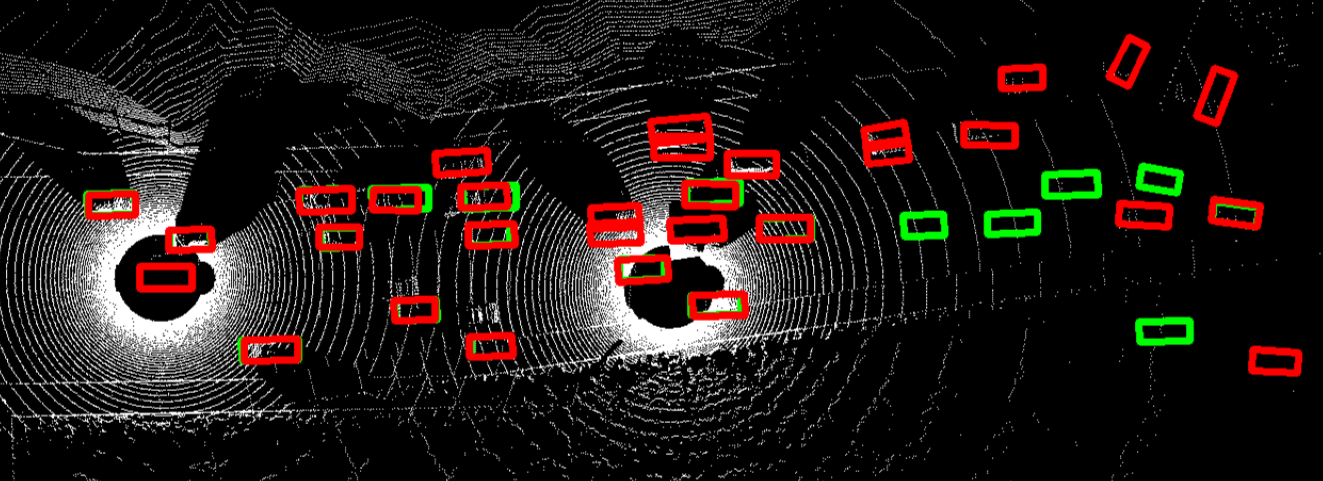}  
  \includegraphics[width=\linewidth]{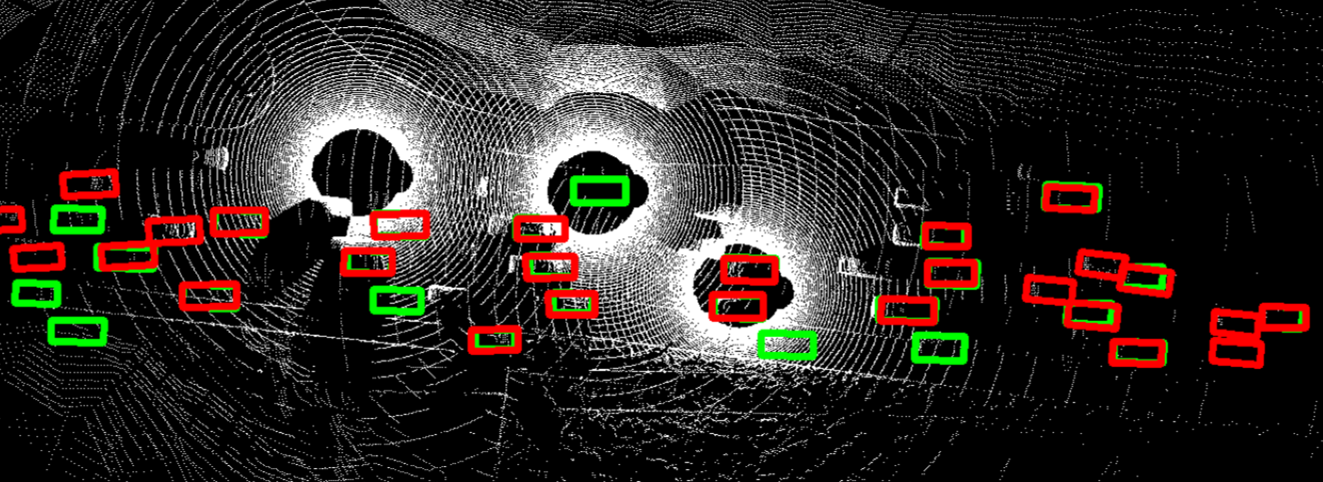}  
  \caption{V2X-ViT}
  \label{subfig:detect_1}
\end{subfigure}
\begin{subfigure}{.32\textwidth}
  \centering
  \includegraphics[width=\linewidth]{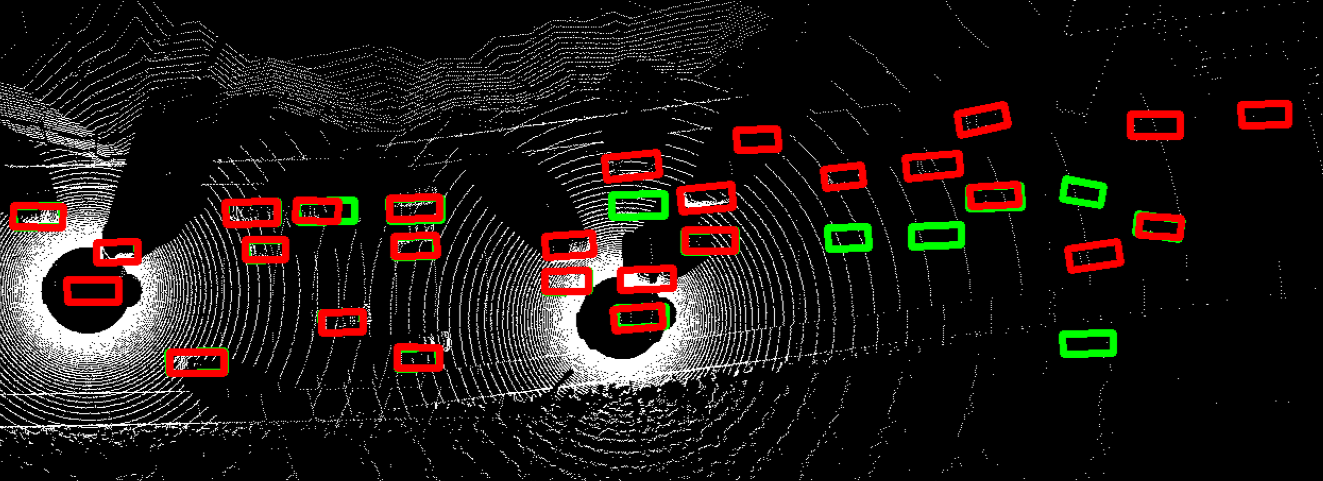} 
  \includegraphics[width=\linewidth]{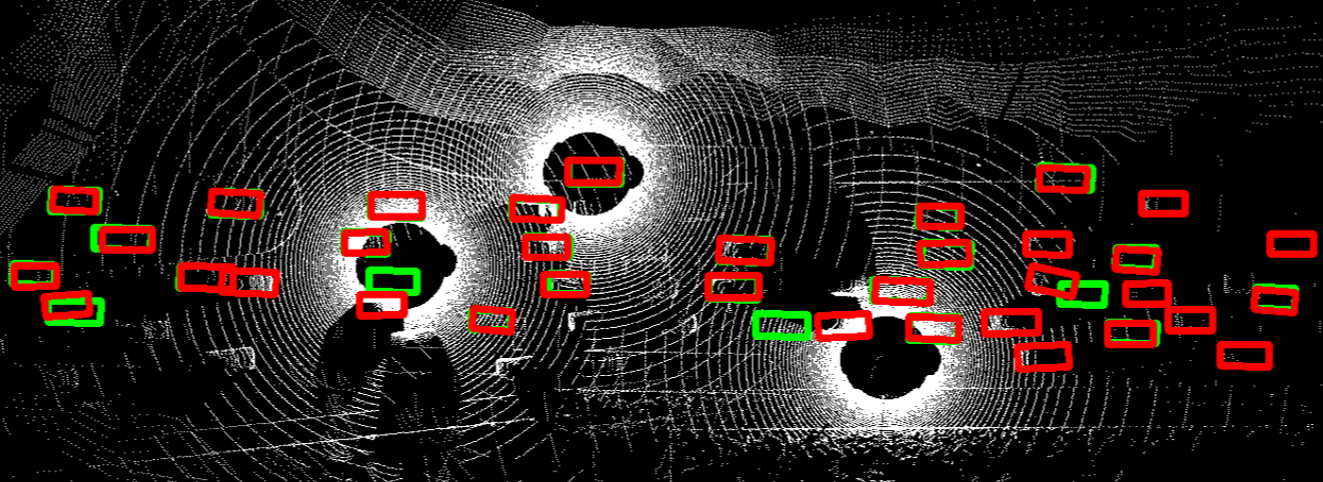} 
  \caption{CoAlign}
  \label{subfig:detect_2}
\end{subfigure}
\begin{subfigure}{.32\textwidth}
  \centering
  \includegraphics[width=\linewidth]{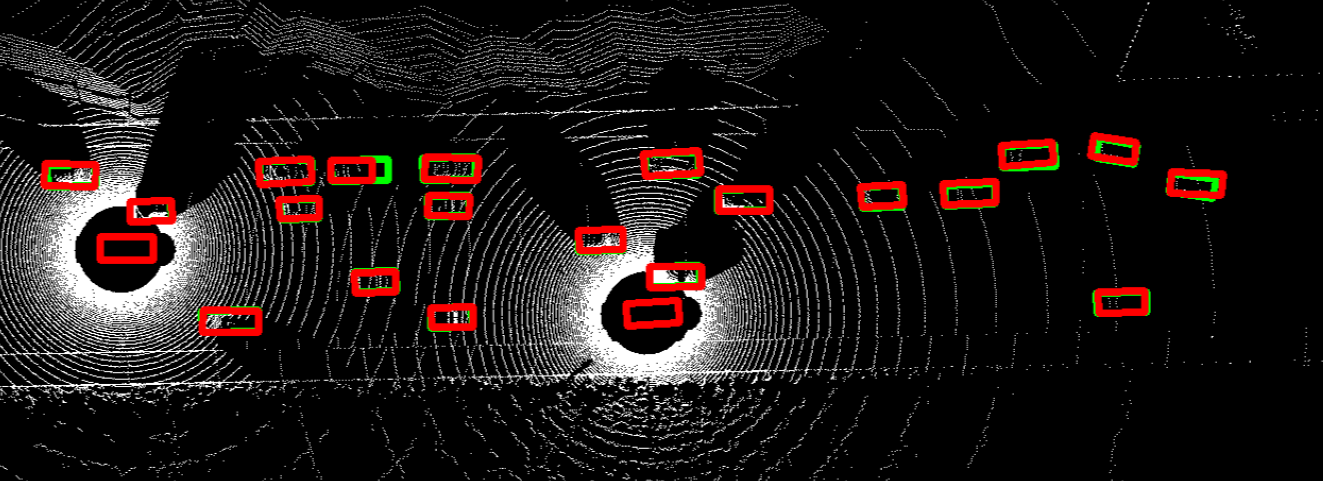}  
  \includegraphics[width=\linewidth]{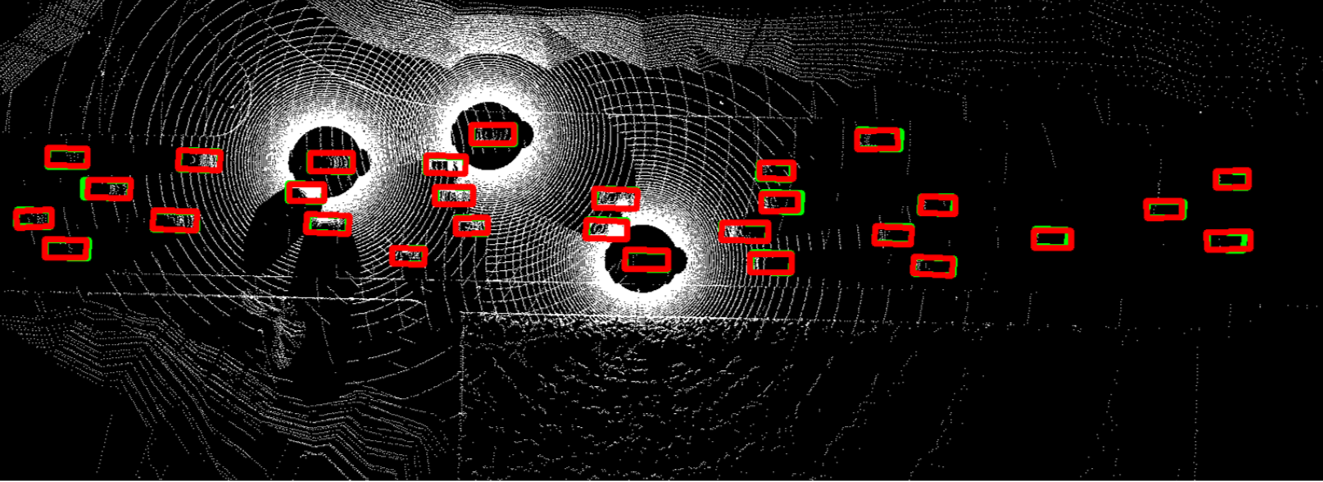} 
  \caption{CoBEVGlue}
  \label{subfig:detect_3}
\end{subfigure}
\caption{\emph{CoBEVGlue} qualitatively outperforms V2X-ViT \cite{xu2022v2x} and CoAlign \cite{lu2023robust} on OPV2V dataset under localization noisy setting. \textcolor{green}{Green} and \textcolor{red}{red} boxes denote ground-truth and detection, respectively. }
\label{fig:detectionresult}
\end{figure*}


\begin{figure*}[t]
\begin{subfigure}{.50\textwidth}
  \centering
  \includegraphics[width=\linewidth]{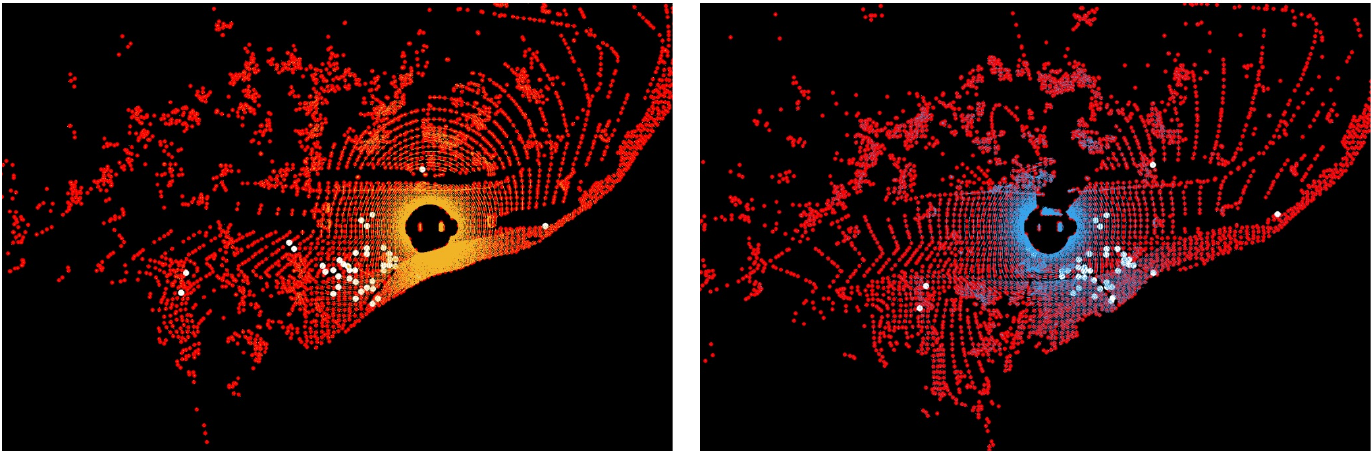}  
  \caption{Point cloud registration}
  \label{subfig:pcr}
\end{subfigure}
\begin{subfigure}{.50\textwidth}
  \centering
  \includegraphics[width=\linewidth]{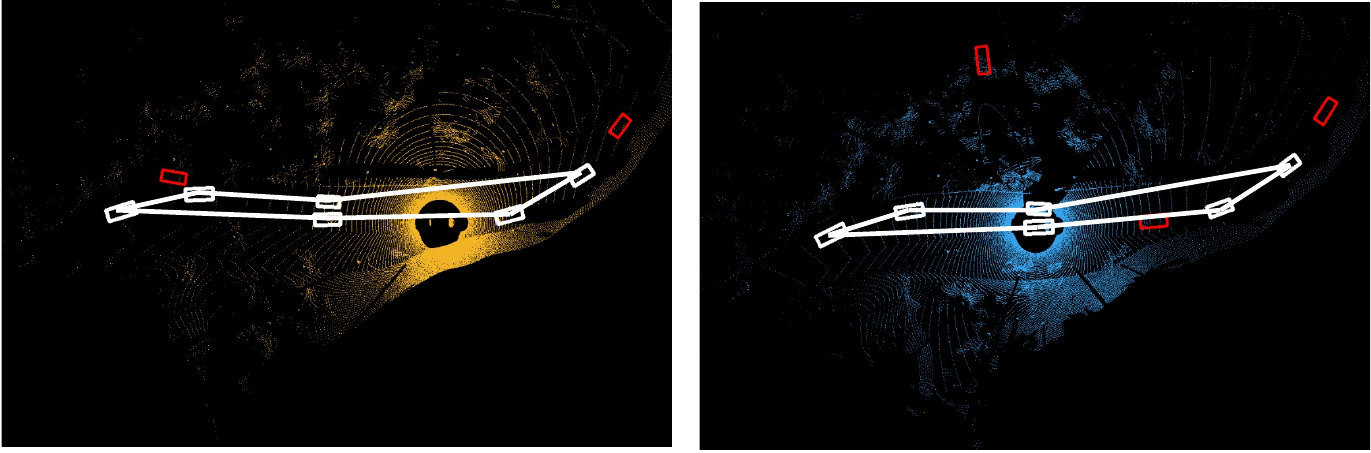}  
  \caption{BEVGlue}
  \label{subfig:bevglue}
\end{subfigure}
\caption{Visualization of calculating relative pose between two collaborators on OPV2V\cite{xu2022opv2v} dataset. \textcolor{yellow}{Yellow} and \textcolor{blue}{blue} denote the point clouds collected from ego vehicle and the collaborator, respectively. In (a), the massive points transmitted between agents are depicted in \textcolor{red}{red} and white, with white specifically indicating the successfully matched points. 
In (b), \textcolor{red}{red} and white represent the boxes transmitted among agents, while white highlights the co-visible objects and their spatial relationships. $\mathtt{BEVGlue}$ achieves accurate relative pose estimation with low communication cost by only transmitting object-wise information. Point cloud registration method takes over 2000 times communication volume than our $\mathtt{BEVGlue}$, making it unusable for bandwidth limited collaborative perception systems. }
\label{fig:qulitative}
\end{figure*}

\subsection{Datasets and Experimental Settings}
We conduct collaborative LiDAR-based 3D object detection on both a simulation dataset, OPV2V \cite{xu2022opv2v}, co-simulated by OpenCDA \cite{opencda} and Carla \cite{carla}, and two real-world dataset, DAIR-V2X \cite{yu2022dair} and V2V4Real\cite{xu2023v2v4real}. We follow \cite{xu2022opv2v,lu2023robust,xu2023v2v4real} to set the detection range as $ x \in [-140m, 140m], y \in [-40m,40m]$ in OPV2V and V2V4Real and $x \in [-100m, 100m], y \in [-40m,40m]$ in DAIR-V2X respectively. We use PointPillars \cite{lang2019pointpillars} with the grid size $(0.4m,0.4m)$ as the encoder. For multi-scale feature fusion, the residual layer number is 3 and the channel numbers are $(64,128,256)$. The communication results count the message size by byte in log scale with base $2$.

\subsection{Quantitative Evaluation}

\textbf{Detection performance in the presence of localization noise.}
Table \ref{table:noise} compares the proposed $\mathtt{CoBEVGlue}$ with previous methods under localization noise on OPV2V, DAIR-V2X and V2V4Real. For the setting of localization noise, we apply Gaussian noise $\mathcal{N}(0,\sigma_t)$ on $x,y$ and $\mathcal{N}(0,\sigma_r)$ on $\theta$, where $x,y, \theta$ are the 2D centers and yaw angle of each agent's accurate 3DoF pose. This noise setting follows previous work \cite{wang2020v2vnet,xu2022v2x,hu2022where2comm,lu2023robust}, while we increase the range of standard deviation to cover more challenging settings. The baseline methods include no collaboration, collaborative methods without specific design to localization noise, and collaborative methods including robust design for localization error. We see that the performances of
$\mathtt{CoBEVGlue}$ are not affected by the level of pose noise since it operates independently of localization systems, significantly outperforming previous methods in various noise levels across both datasets.  

\textbf{Detection performance in the presence of localization attack.}
We also explore detection performance under the common and unsolved GPS Spoofing attack \cite{zeng2018all,bhatti2017hostile}, an attack where malicious attackers set arbitrary position by sending fake satellite
signals. Specifically, we consider the attacker to deceive all collaborators into thinking they are in the same location, aiming to generate false positive bounding boxes. Table~\ref{tab:attack} presents the detection performance and communication bandwidth of various methods under this attack, along with their performance when integrated with our spatial alignment module $\mathtt{BEVGlue}$. The results reveal that $\mathtt{BEVGlue}$ significantly improves performance under attack while bringing negligible communication overhead.
Notably, with $\mathtt{BEVGlue}$'s assistance, a majority of collaborative perception methods outperform single-agent perception even under malicious attack.


\textbf{Comparisons with point cloud registration.}
We present a comparison between $\mathtt{BEVGlue}$ and two representative point cloud registration methods in Fig.~\ref{fig:icp}. This comparison includes both 2D and 3D versions of the Iterative Closest Point (ICP) and a widely used pipeline in recent multi-agent systems \cite{lajoie2023swarm}, which incorporates Fast Point Feature Histograms (FPFH) \cite{rusu2009fast} for keypoint description and TEASER++ \cite{teaser} for robust registration. Initial poses provided to ICP are varied under different levels of Gaussian noise. We see that i) $\mathtt{CoBEVGlue}$ consistently delivers superior performance with minimal communication volume, and its effectiveness remains stable regardless of the extent of localization noise;  ii) while ICP can mitigate the impact of minor localization noise, it consumes large communication bandwidth and its performance significantly declines under bandwidth constraints; iii) as localization noise increases, ICP fails to achieve successful alignment, irrespective of the available bandwidth.

\textbf{Computation time.} Tested on a system equipped with a 2.90GHz Intel Xeon CPU and an RTX 4090 GPU, $\mathtt{BEVGlue}$ achieves 89.98 frames per second (FPS) on OPV2V, 72.18 FPS on DAIR-V2X and 158.7 FPS on V2V4Real.

\subsection{Qualitative Evaluation}

\textbf{Visualization of detection results.}
Fig.~\ref{fig:detectionresult} shows a comparative visualization of detection results from V2X-ViT, CoAlign, and $\mathtt{CoBEVGlue}$ in the OPV2V dataset under noisy setting. The noise stems from a Gaussian distribution with a standard deviation of 3.0m for position and 3.0° for heading. V2X-ViT, despite employing the MSWin module to mitigate pose error, struggles under large noise. Similarly, the pose graph optimization algorithm in CoAlign fails in the presence of large noise, leading to a severe drop in detection performance. In contrast, $\mathtt{CoBEVGlue}$'s exhibits superior performance under large noise. This can be attributed to its independence from prior pose information, which makes it less susceptible to the impacts of pose noise.

\textbf{Visual comparison with point cloud registration.}  
Fig.~\ref{fig:qulitative} compares the collaborative perception system with point cloud registration and the one with $\mathtt{BEVGlue}$. For point cloud registration, we use FPFH \cite{rusu2009fast} for feature descriptor and TEASER \cite{teaser} for registration. Fig.~\ref{subfig:pcr} shows that the point cloud registration pipeline identifies a few matched points (in white) from a large amount of transmitted points; and Fig.~\ref{subfig:bevglue} shows that $\mathtt{BEVGlue}$ uses the spatial geometry of objects to find correspondence. Compared with the point cloud registration pipeline, $\mathtt{BEVGlue}$ uses much less communication volume ($2000\times$).

\subsection{Ablation studies}


\begin{table*}[t]
\centering
\caption{Ablation study.} \vspace{-1mm}
\resizebox{\textwidth}{!}{

\begin{tabular}{cccccccc}
\hline
\multicolumn{2}{c}{Dataset}                       & \multicolumn{3}{c}{OPV2V}                                                & \multicolumn{3}{c}{V2V4Real}                        \\ \hline
\multicolumn{2}{c|}{Matching Criteria}            & \multicolumn{6}{c}{Metric}                                                                                                     \\ \hline
Geometric pattern & \multicolumn{1}{c|}{Tracking} & AP@0.3$\uparrow$          & AP@0.5$\uparrow$          & \multicolumn{1}{c|}{AP@0.7$\uparrow$}          & AP@0.3$\uparrow$          & AP@0.5$\uparrow$          & AP@0.7$\uparrow$          \\ \hline
                  & \multicolumn{1}{c|}{}         & 0.7904          & 0.7781          & \multicolumn{1}{c|}{0.6746}          & 0.4919          & 0.4469          & 0.2611          \\
\checkmark                & \multicolumn{1}{c|}{}         & 0.9614          & 0.9545          & \multicolumn{1}{c|}{0.9049}          & 0.6839          & 0.6356          & 0.3787          \\
\checkmark                & \multicolumn{1}{c|}{\checkmark }      & \textbf{0.9651} & \textbf{0.9581} & \multicolumn{1}{c|}{\textbf{0.9088}} & \textbf{0.7394} & \textbf{0.7016} & \textbf{0.4308} \\ \hline
\end{tabular}
\vspace{-3mm}
}
\label{tab:ablation}
\end{table*}

Table \ref{tab:ablation} assesses the effectiveness of the geometric pattern and tracking information used in maximum common subgraph detection on the OPV2V and V2V4Real dataset. Absent the geometric pattern, the matching process only relies on node information, ignoring the spatial relationship between nodes. Without the tracking information, the search ignores temporal consistency. We see that: 1) Geometric patterns play a crucial role in enhancing our method for common subgraph detection. Specifically, on the V2V4Real dataset, integrating geometric patterns results in a performance boost of 45.7\% over the baseline; and 2) The inclusion of tracking information significantly contributes to the temporal coherence of the matching outcomes, with an improvement of 57.0\% on the V2V4Real dataset compared to the baseline approach.

\section{Conclusions}
This paper proposes a novel self-localized collaborative perception framework $\mathtt{CoBEVGlue}$ and a novel spatial alignment method $\mathtt{BEVGlue}$. The core idea is to search for co-visible objects from the bird’s eye view perceptual data across agents and calculate the relative pose between agents, ensuring a consistent spatial coordinate system for collaboration. Comprehensive experiments show that $\mathtt{CoBEVGlue}$ performs comparably to systems relying on precise localization information and achieves state-of-the-art detection performance when localization noise and attack exist.

\textbf{Limitation and future work.} This work focuses on exploiting spatial alignment for collaborative perception. We plan to mitigate the temporal misalignment issue in future.
%
%
\bibliographystyle{splncs04}
\bibliography{main}
\end{document}